\documentclass{article}
\usepackage{spconf,amsmath,graphicx}
\usepackage{amsfonts}
\usepackage{booktabs}
\usepackage{multirow}
\usepackage{bm}
\usepackage{amssymb}
\usepackage[linesnumbered, lined, ruled]{algorithm2e}
\usepackage{microtype}
\usepackage{bbm}
\usepackage{cite}

\title{DRBANet: A lightweight dual-resolution network for semantic \\ segmentation with Boundary auxiliary}
\name{Linjie Wang$^{1}$ , Quan Zhou$^{1, *}$\thanks{Corresponding author: Quan Zhou, quan.zhou@njupt.edu.cn,This work is partly supported by NSFC (No. 61876093), NSFJS (No. BK20181393), and NSF (No. IIS-1302164).}, Chenfeng Jiang$^{1}$, Xiaofu Wu$^{1}$, and Longin Jan Latecki$^{2}$
}
\address{$^{1}$National Engineering Research Center of Communications and Networking, \\ Nanjing University of Posts \& Telecommunications, P.R. China.\\
 $^{2}$Department of Computer and Information Sciences, Temple University, Philadelphia, USA.\\}
 
\begin{document}
\maketitle

\begin{abstract}
Due to the powerful ability to encode image details and semantics, many lightweight dual-resolution networks have been proposed in recent years. However, most of them ignore the benefit of boundary information. This paper introduces a lightweight dual-resolution network, called DRBANet, aiming to refine semantic segmentation results with the aid of boundary information. DRBANet adopts dual parallel architecture, including: high resolution branch (HRB) and low resolution branch (LRB). Specifically, HRB mainly consists of a set of Efficient Inverted Bottleneck Modules (EIBMs), which learn feature representations with larger receptive fields. LRB is composed of a series of EIBMs and an Extremely Lightweight Pyramid Pooling Module (ELPPM), where ELPPM is utilized to capture multi-scale context through hierarchical residual connections. Finally, a boundary supervision head is designed to capture object boundaries in HRB. Extensive experiments on Cityscapes and CamVid datasets demonstrate that our method achieves promising trade-off between segmentation accuracy and running efficiency.
\end{abstract}


\section{Introduction}\label{sec:intro}

\begin{figure*}[ht] 
\centering 
\includegraphics[width=1.0\textwidth]{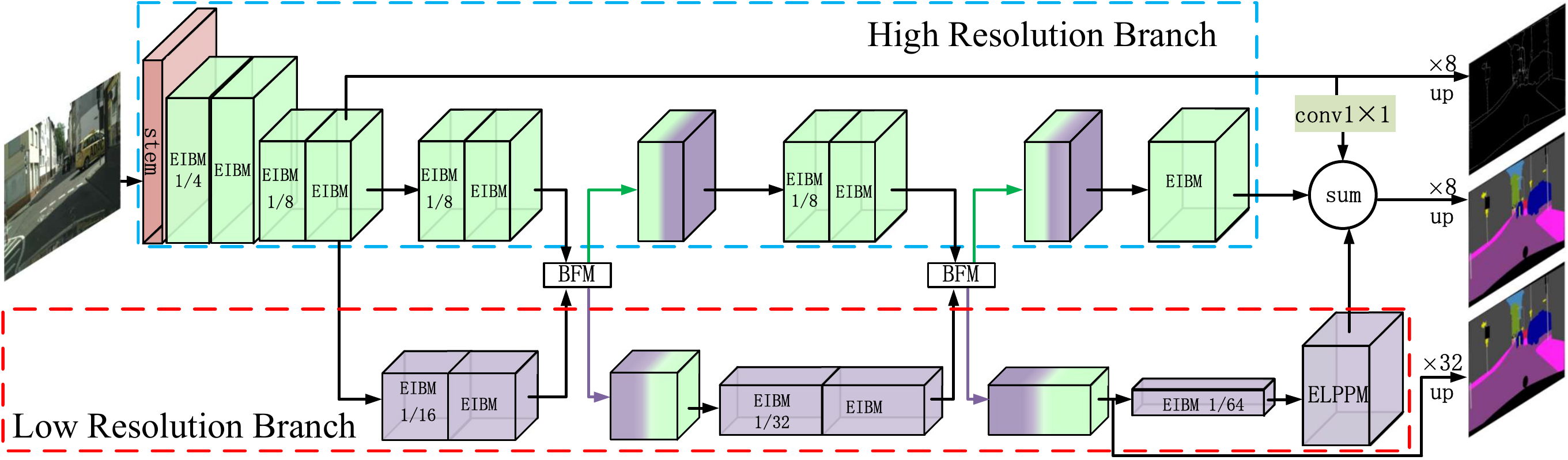}
\vspace{-0.7cm}
\caption{The overall architecture of DRBANet. HRB and LRB are denoted by blue and red dash bounding boxes, respectively. Green and purple arrows indicate integrated information flow in two branches. (Best viewed in color)} 
\label{fig:architecture} 
\end{figure*}

Semantic segmentation plays a significant role in some real-world applications, such as augmented reality, robot sensing, autonomous driving, and so on. The goal of semantic segmentation is to assign a unique semantic category label to each pixel in image. With the development of convolutional neural networks (CNNs), some accurate networks \cite{fcn,ocrnet,deeplab,pspnet,ccnet} have been proposed for semantic segmentation, which have hundreds even thousands of convolutional layers and feature channels. Due to their complicated network architecture, it is difficult to deploy them into resource-constrained edge devices.

To achieve online estimation in timely fashion, many researchers prefer to design lightweight semantic segmentation networks \cite{hyperseg,espnet,espnetv2,erfnet,dabnet,stdcnet,bisenet,bisenetv2,fast-scnn,lite-hrnet,gas}, which can be roughly classified into two categories: single path networks \cite{espnet,espnetv2,erfnet,dabnet,stdcnet} and dual-resolution networks \cite{bisenet,bisenetv2,fast-scnn}. The first category often designs lightweight backbone to extract features. For example, ERFNet \cite{erfnet} utilizes decomposition convolution to remain accuracy and reduce model size. ESPNetV2 \cite{espnetv2} proposes extremely efficient spatial pyramid unit to enlarge receptive fields. DABNet \cite{dabnet} employs depthwise asymmetric bottleneck to capture local context. STDCNet \cite{stdcnet} designs short-term dense concatenate module to obtain scale-variant receptive fields. On the other hand, the second category usually employs compact dual-resolution architecture for semantic segmentation, where low resolution is used to capture high-level semantics, while high resolution is designed to remain fine image details. For instance, BiSeNet \cite{bisenet} divides the network into spatial and context paths separately, where both of them involve lightweight architecture. BiSeNetV2 \cite{bisenetv2}, as an extension of \cite{bisenet}, proposes a finer way to fuse features from two branches, leading to great reduction of model size. In \cite{lite-hrnet}, an $1 \times 1$ convolution is replaced by cross-resolution weighting module in HRNet \cite{hrnet}. In spite of achieving impressive performance, both categories neglect the benefit of boundary information, which provides additional cues to boost performance. Moreover, most of these networks merely capture context cues in one single scale, which is always not enough to make final discrimination for each individual pixel.

To deal with these shortcomings, this paper designs a lightweight dual-resolution network, named DRBANet,  for semantic segmentation with boundary auxiliary. DRBANet adopts dual parallel architecture, including: high resolution branch (HRB) and low resolution branch (LRB). Specifically, HRB is designed for remaining high resolution spatial details, while LRB is used to capture high level semantic features via fast downsampling strategy. As shown in Fig. \ref{fig:architecture}, DRBANet includes four components: stem, Efficient Inverted Bottleneck Module (EIBM), Extremely Lightweight Pyramid Pooling Module (ELPPM), and Bilateral Fusion Module (BFM). As the main unit of DRBANet, EIBM employs inverted bottleneck structure \cite{mobilenetv2} with multiple depthwise convolution layers to enlarge receptive fields. ELPPM is designed at the end of LRB to further capture multi-scale semantic context. Unlike previous methods \cite{bisenet,bisenetv2} that only extract convolution features in two branches independently, BFM integrates features with different resolutions to enhance information communication between HRB and LRB. To fully explore the detail cues in HRB, a boundary supervision head is used at the top of DRBANet to extract object boundary cues. Finally, the features, calculated from dual-resolution branches and boundary head, are fused together to predict final semantic outputs. In summary, the contributions of this paper are three-fold: (1) The dual-resolution structure of DRBANet leverages network size and feature representation. (2) ELPPM extracts multi-scale semantic context without considerable increase of computational complexity. (3) Boundary information is used as additional auxiliary to improve semantic segmentation.

\section{OUR METHOD}\label{sec:method}
\subsection{Network Architecture}\label{ssec:architecture}

DRBANet follows a dual-branch architecture \cite{bisenet,bisenetv2}, where HRB produces image detail features, while LRB captures image semantic cues. DRBANet is built mainly based on EIBM unit, which enables us to explore larger receptive fields, but with very smaller computational overhead. To enhance representation capability, BFMs are also repeatedly used as bridges to enable communications between HRB and LRB. The main network architecture is depicted in Tab. \ref{tab:DRBANet}. The HRB consists of layers from 1 to 12, where the first layer is a stem, and the rests are EIBMs. The stem layer utilizes stride $3 \times 3$ convolution to reduce feature resolution. Thereafter, the feature size is reduced twice, resulting in the resolution of $\frac{1}{4}$ and $\frac{1}{8}$ with respect to input image. On the other hand, layers from 6 to 13 form LRB, composed by EIBMs and ELPPM. In this path, the feature size is sequentially downsampled via EIBM, leading to the resolution of $\frac{1}{16}$, $\frac{1}{32}$, and $\frac{1}{64}$ of input image. At the end of LRB, ELPPM encodes multi-scale context and recovers equal feature dimensions to the output of HRB. A boundary supervision head is added at layer 5, where the associated features undergo an $1\times1$ convolution, resulting in equal resolutions with the outputs of HRB and LRB for following fusion. At the top of DRBANet, we have one estimated boundary map, and two predicted semantic maps (upsampled $8\times $, $8\times $, and $32 \times$, from layer 5, fused features, and layer 12 of LRB), receiving their supervisions from the corresponding ground truth.

\subsection{EIBM and BFM}\label{ssec:EIBM_BFM}
\begin{table}[!t]
\vspace{-0.2cm}
\small \tabcolsep 0.5pt \caption{The architecture of DRBANet. ``Size'' denotes the dimension of output feature maps, $s$ denotes stride 2.}
\begin{center}
\scalebox{0.85}{
\begin{tabular}{l|c|l|l}
\toprule
\textbf{Layer}      &\textbf{Size}        &\multicolumn{2}{l}{\textbf{\quad\quad~ DRBANet}} \\
\midrule
$\begin{array}{l}1\end{array}$         &$\begin{array}{l}512\times512\times32\end{array}$ &\multicolumn{2}{c}{~~~ stem ($3\times 3$, s)} \\	
\midrule
$\begin{array}{l}2-3\\4-5\end{array}$    &$\begin{array}{l}256\times256\times32\\128\times128\times64\end{array}$  &\multicolumn{2}{c}{$\left[\begin{array}{l} \mathrm{EIBM,s}\\ \mathrm{EIBM}\end{array}\right] \times 2 $}\\
\midrule
$\begin{array}{l}6-8\\9-11 \end{array}$    &$\begin{array}{l}64\times64\times128,128\times128\times64\\32\times32\times256,128\times128\times64 \end{array}$  &\multicolumn{2}{l}{$\left[ \begin{array}{l|l} \mathrm{EIBM,s}&\mathrm{EIBM} \\\mathrm{EIBM} & \mathrm{EIBM}\\ \hline \multicolumn{2}{c}{\mathrm{BFM}} \end{array}\right] \times 2 $}\\
\midrule
$\begin {array}{l}12\\13\end{array}$  &$\begin{array}{l}\ 16 \ \times\ 16\times512,128\times128\times128\\ 128\times128\times128, ~~~~~~~~~~-~~\end{array}$     &\multicolumn{2}{l}{$ \begin{array}{l|l} \mathrm{\;\; EIBM,s}&\mathrm{EIBM} \\\mathrm{\;\; ELPPM} &\;\;~~-~~\end{array}$}\\
\bottomrule
\end{tabular}}
\end{center}
\label{tab:DRBANet}
\end{table}

\begin{figure}[ht] 
\centering 
\includegraphics[scale=0.9]{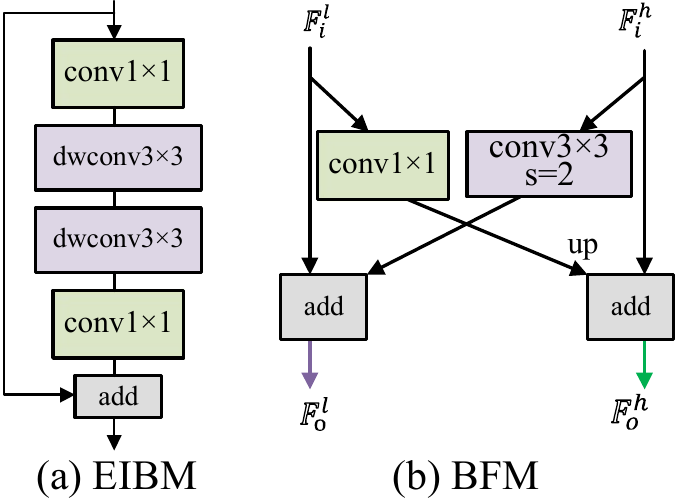}
\vspace{-0.45cm}
\caption{ The detail of EIBM (a) and BFM (b). The arrows with different colors denote corresponding information flow in Fig. \ref{fig:architecture}. Note second BFM utilizes two successive $3 \times 3$ convolutions with stride 2 to reduce resolution. (Best viewed in color)} 
\label{fig:EIBM_BFM} 
\vspace{-0.2cm}
\end{figure}
As shown in Fig. \ref{fig:EIBM_BFM}(a), EIBM is designed based on inverted bottleneck block. At the beginning of EIBM, the channel number of input is expanded $2 \times$ via an $1 \times 1$ convolution. To enlarge receptive fields, two convolution layers are involved using $3\times 3$ depthwise convolution, resulting in the independent filtering responses among all feature channels. Thereafter, an $1\times1$ convolution is used to recover channel dependencies using a linear combination. Finally, a skipped-connection is employed to leverage lightweight convolution and end-to-end training. Note EIBM can be also used to reduce feature resolutions, accomplished using  stride depthwise separable convolution, as shown in Tab. \ref{tab:DRBANet}.

As shown in Fig. \ref{fig:EIBM_BFM}(b), BFM is used to enhance feature communication between HRB and LRB. Let $\mathbb{F}_{i}^{l} \in \mathbb{R}^{H \times W \times C}$ and $\mathbb{F}_{i}^{h} \in \mathbb{R}^{H^{\prime} \times W^{\prime} \times C^{\prime}}$ be inputs of BFM, and $\mathbb{F}_{o}^{l} \in \mathbb{R}^{H \times W \times C}$ and $\mathbb{F}_{o}^{h} \in \mathbb{R}^{H^{\prime}  \times W^{\prime}  \times C^{\prime} }$ be outputs of BFM, respectively.  $\mathbb{F}_{i}^{l}$ first passes through an $1 \times 1$ convolution $\textbf{\emph{f}}_{1\times1}$, and then upsampled with equal dimensions for following feature fusion with $\mathbb{F}_{i}^{h}$. On the other hand, to produce $\mathbb{F}_{o}^{l}$, $\mathbb{F}_{i}^{h}$ is directly fed into a $3 \times 3$ stride convolution $\textbf{\emph{f}}_{3\times3}$, and then integrated with $\mathbb{F}_{i}^{l}$. Actually, two types of convolutions can be considered as a cross-resolution residual function, which is helpful to train BFM in an end-to-end manner. 
\vspace{-0.2cm}
\begin{equation}
        \mathbb{F}_{o}^{h}=\mathbb{F}_{i}^{h} \oplus \textbf{U}(\mathbb{F}_{i}^{l} \ast \textbf{\emph{f}}_{1\times1}) \hspace{10pt}
        \mathbb{F}_{o}^{l}=\mathbb{F}_{i}^{l} \oplus (\mathbb{F}_{i}^{h}\ast \textbf{\emph{f}}_{3\times3})
\vspace{-0.2cm}
\end{equation}
where $\ast$ indicates convolution operation, $\oplus$ denotes element-wise addition, and $\textbf{U}(\cdot)$ stands for upsampling.

\subsection{ELPPM}\label{ssec:ELPPM}

\begin{figure}[ht] 
\centering 
\includegraphics[scale=0.65]{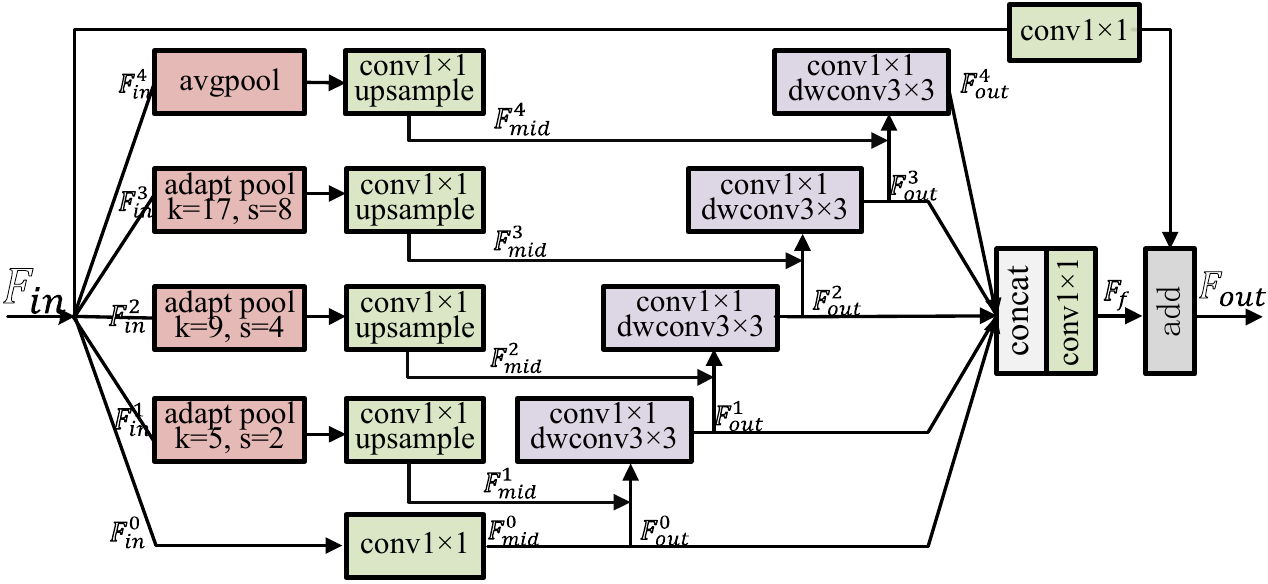}
\vspace{-0.2cm}
\caption{The detail of ELPPM. (Best viewed in color)} 
\label{fig:ELPPM} 
\vspace{-0.2cm}
\end{figure}

As illustrated in Fig. \ref{fig:ELPPM}, this section introduces ELPPM that captures multi-scale context of LRB. Motivated from \cite{res2net}, ELPPM utilizes hierarchical residual-like connections to blend information from different receptive fields, but with more lightweight structure. The input is first fed into five parallel paths $\mathbb{F}_{in}^{i} \in \mathbb{R}^{H\times W\times C}$, ${i} \in \{0,1,..4\}$. Except $\mathbb{F}_{in}^{4}$ that performs global average pooling $\textbf{P}_g(\cdot)$, other three paths employ adaptive stride pooling $\textbf{P}_i(\cdot)$, $i \in \{1, 2, 3\}$ with kernel size \{5, 9, 17\}, resulting in the fact that the size of pooling features are sequentially reduced. Thereafter, an $1\times1$ convolution $\textbf{\emph{f}}_{1\times1}$ is used to reduce channel numbers from $C$ to $C/4$. Finally, except $\mathbb{F}_{in}^{0}$, upsampling operation $\textbf{U}(\cdot)$ is utilized in each path to produce $\mathbb{F}_{mid}^{i} \in \mathbb{R}^{H\times W\times C/4}$ for following fusion. 
\vspace{-0.2cm}
\begin{equation}
\mathbb{F}_{mid}^{i} = \begin{cases}
\textbf{\emph{f}}_{1\times1} \ast \mathbb{F}_{in}^{i}, &i=0;\\
\textbf{U}(\textbf{\emph{f}}_{1\times1} \ast \textbf{P}_i(\mathbb{F}_{in}^{i})), &0<i<4; \\
\textbf{U}(\textbf{\emph{f}}_{1\times1} \ast \textbf{P}_{g}(\mathbb{F}_{in}^{i})), &i=4.  \\
\end{cases}
\vspace{-0.2cm}
\end{equation}

To obtain the output $\mathbb{F}_{out}^{i} \in \mathbb{R}^{H\times W\times C/4}$ in $i^{th}$ path, $\mathbb{F}_{mid}^{i}$ and $\mathbb{F}_{out}^{i-1}$ are combined by addition. Then, a depthwise convolution and an $1 \times 1$ point convolution are employed to integrate information in neighbor paths. Note there is no convolution in the first path, since $\mathbb{F}_{mid}^{0}$ is directly used as $\mathbb{F}_{out}^{0}$. In spite of having very lightweight structure, the parallel paths can be considered as a group convolution \cite{xie2017agg}, as there are no relationship among all paths. To recover channel dependencies and reduce dimension, output features $\mathbb{F}_{out}^{i}$ in all paths are stacked together, and fed into an $1 \times 1$ convolution, producing fused features $\mathbb{F}_{f} \in \mathbb{R}^{H\times W\times C/4}$.

Finally, ELPPM leverages multi-path convolution and residual connections in an end-to-end training manner. Specifically, the input
$\mathbb{F}_{in}$ firstly undergoes an $1 \times 1$ convolution to compress channel numbers, and then added with $\mathbb{F}_{f}$, generating output $\mathbb{F}_{out}$ of ELPPM. Note the size of $\mathbb{F}_{out}$ has to be enlarged 8 $\times$ for integration with other parts of DRBANet.



\subsection{BSH}\label{ssec:BSH}

\begin{figure*}[ht] 
\centering 
\includegraphics[width=1\textwidth]{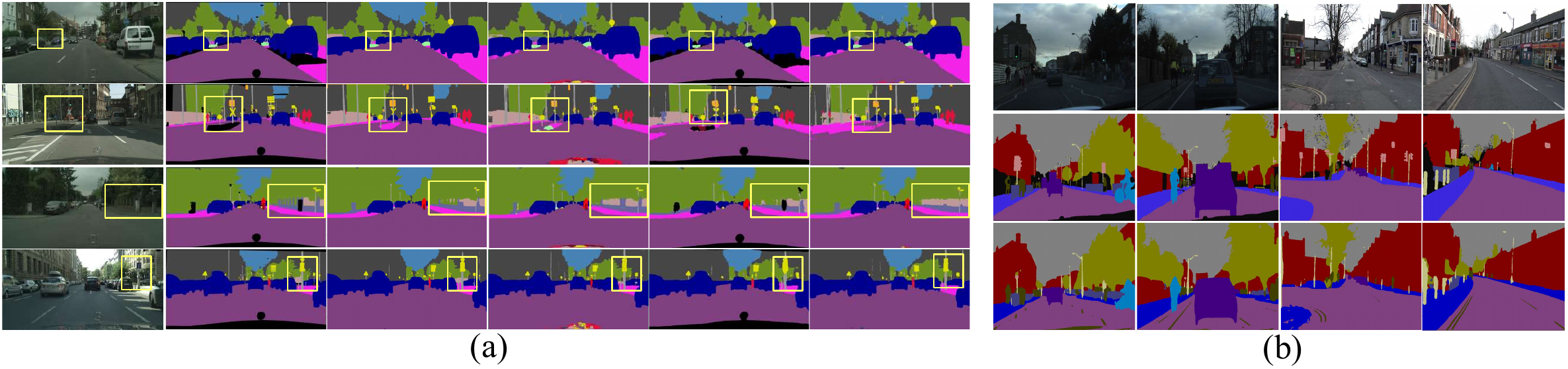}
\vspace{-1cm}
\caption{Comparison of some visual examples of semantic segmentation on Cityscapes validation set (a) and CamVid test set (b). In (a), from left to right are images, ground truth, segmentation outputs from DRBANet, DABNet \cite{dabnet}, ESPNetV2 \cite{espnetv2}, and ERFNet \cite{erfnet}. In (b), from top to down are images, ground truth, and outputs of DRBANet. (Best viewed in color)} 
\label{fig:seg_test} 
\end{figure*}

This section designs a boundary supervision head to refine semantic segmentation results. Give the ground truth of semantic segmentation, we adopt Laplacian kernel to obtain boundary ground truth. Specifically, a series of binary boundary features are produced by Laplacian operator with different strides on semantic segmentation ground truth. Then the features are upsampled to original size and fed into a trainable $1 \times 1$ convolution for fusing multi-scale boundary information. Finally, a threshold 0.1 was utilized to convert the fused feature into boundary ground truth. Considering the class imbalance problem between boundary and non-boundary pixels, we use the binary cross-entropy loss $L_{bce}$ and the dice loss $L_{dice}$ together, where the final boundary loss $L_{bound}$ can be written as:
\vspace{-0.2cm}
\begin{equation}
L_{bound}(\emph{\textbf{p}}, \emph{\textbf{g}}) = L_{bce}(\emph{\textbf{p}}, \emph{\textbf{g}})+L_{dice}(\emph{\textbf{p}}, \emph{\textbf{g}})
\label{eq:boundloss}
\vspace{-0.2cm}
\end{equation}
where $\emph{\textbf{p}}$ and $\emph{\textbf{g}}$ stand for boundary predictions and produced ground truth, respectively. 

\section{EXPERIMENTS}\label{sec:experiments}

\subsection{Implementation Details}\label{ssec:subhead}

{\bf Dataset.} We evaluate our network on Cityscapes \cite{cityscapes} and CamVid \cite{camvid} datasets. Cityscapes contains 5,000 pixel-wise annotated images with $2048 \times 1024$ image resolution, where 2,975 for train, 500 for val and 1,525 for test. CamVid consists of 701 densely annotated frames with $960 \times 720$ image resolution, in which 367 for train, 101 for val and 233 for test.

\noindent{\bf Parameter settings.} DRBANet is implemented in the hardware platform of the deep learning server with RTX 2080Ti GPU. We use SGD  algorithm to optimize DRBANet, where momentum and weight decay are set to 0.9 and $5\times e^{-4}$. The batch size is set as 16 and 4 for Cityscapes and CamVid datasets, respectively. The initial learning rate is set to $10^{-2}$ and $10^{-3}$, and the cosine learning policy \cite{consin} is adopted to train our model within 120k, 20k iterations for Cityscapes and CamVid datasets, respectively. Additionally, images are randomly cropped into $1024 \times 1024$ for Cityscapes.

\noindent{\bf Loss settings.} In Fig. \ref{fig:architecture}, there are three supervisions: auxiliary \textls[-25]{boundary loss $L_{bound}$, auxiliary segmentation loss $L_{auseg}$, and} segmentation loss $L_{seg}$. Except first one, the rests adopt cross-entropy loss\cite{fcn,ocrnet,deeplab,pspnet,ccnet}, thus the total loss $L_{total}$ is defined as:
\vspace{-0.2cm}
\begin{equation}
        L_{total}= L_{seg} +\lambda_{1}\times L_{auseg} + \lambda_{2}\times L_{bound}
\vspace{-0.2cm}
\label{eq:total_loss}
\end{equation}
where $L_{bound}$ is defined in Eqn. (\ref{eq:boundloss}), and the  non-negative parameters $\lambda_{1}$ and $\lambda_{2}$ are set to 0.2 and 0.1, empirically.

\subsection{Evaluation Results}

\newcommand{\tabincell}[2]{\begin{tabular}{@{}#1@{}}#2\end{tabular}}
\begin{table}
\vspace{-0.2cm}
\caption{Comparison with other approaches. `-/-' represents the results on test set of Cityscapes and CamVid datasets, respectively. `${*}$' means trained by train and val set together.}
\vspace{-0.2cm}
\center
\begin{tabular}{l|cccc}
\hline  
\noalign{\smallskip}
\multirow{2}{*}{Method} &\multirow{2}{*}{\tabincell{c}{Flops\\(G)}} & \multirow{2}{*}{\tabincell{c}{Params\\(M)}} & \multicolumn{2}{c}{mIoU(\%)} \\
\cline{4-5}\noalign{\smallskip}  &   &  & val &test \\ 
\noalign{\smallskip}\hline\noalign{\smallskip}
 ERFNet \cite{erfnet} &27.7     &20     &70.0   &68.0/~~~-~~~  \\
  BiSeNetV1 \cite{bisenet} &2.9      &5.8   &69.0      &68.4/65.6 \\
  ESPNetV2 \cite{espnetv2} &2.7       &~~-~~~        &66.4       &66.2/~~~-~~~ \\
  DABNet \cite{dabnet}&~~-~~~&0.8&~~-~~~&70.1/66.4\\
  FasterSeg \cite{fasterseg} &28.2&4.4 &73.1&71.5/71.1  \\
  BiSeNetV2 \cite{bisenetv2} &21.2&~~-~~~&73.4& 72.6/\textbf{76.7}  \\
  STDCNet \cite{stdcnet} &~~-~~~&~~-~~~&74.2&73.4/73.9 \\
  \noalign{\smallskip}\hline\noalign{\smallskip}
  Ours       &11.9    &2.3  &\textbf{76.6}    &74.3/72.6 \\  
  Ours$^{*}$ &11.9    &2.3  &~~-~~~   &\textbf{75.1}/73.9\\
  \noalign{\smallskip}\hline\noalign{\smallskip}
 \end{tabular}
 \label{tab:seg_miou}
\end{table}

\noindent{\bf Results on Cityscapes.}\label{ssec:cityscapes}
Tab. \ref{tab:seg_miou} reports comparison results with selected state-of-the-art networks, demonstrating that our DRBANet achieves the best trade-off between accuracy and efficiency. From Tab. \ref{tab:seg_miou}, DRBANet achieves 75.1\% mIoU on test set, with only 11.9 GFLOPs and 2.3M model size. Moreover, our network has nearly $2\times$ less parameters and $2.3\times$ smaller GFlops than FasterSeg \cite{fasterseg}, and improves segmentation accuracy with 3.6\% mIoU. Although BiSeNetV1 \cite{bisenet}, another efficient network, is nearly $4\times$ efficient, but has $2.5\times$ larger model size and delivers poor segmentation accuracy of 6.7\% mIoU drop than DRBANet on Cityscapes test set. Fig. \ref{fig:seg_test}(a) shows some visual examples of segmentation outputs on the Cityscapes val set. It is demonstrated that DRBANet produces more consistent visual outputs with accurate and delineated object boundaries (denoted by yellow bounding boxes).

\noindent{\bf Results on CamVid.}\label{ssec:camvid}
We also evaluate our method on CamVid dataset. As shown in Tab. \ref{tab:seg_miou}, DRBANet obtains 73.9\% mIoU, achieving a good trade-off between performance and efficiency. Compared with BiSeNetV2 \cite{bisenetv2}, DRBANet has 2.8\% mIoU drop, yet it only requires nearly half GFLOPs with respect to \cite{bisenetv2}. To further demonstrate the superior performance of our method, several visual examples of qualitative segmentation outputs are shown in Fig. \ref{fig:seg_test}(b). DRBANet is able to accurately segment tiny object such as ``pole'', and correctly identify object within different scales, such as ``pedestrian''.

\section{CONCLUSION Remarks AND FUTURE WORK}
\label{sec:conclusion}

This paper has designed a lightweight DRBANet for semantic segmentation with boundary auxiliary. DRBANet leverages segmentation accuracy and implementing efficiency mainly from EIBM and ELPPM. EIBM utilizes two sequential depthwise convolution layers to enlarge receptive fields but at a smaller computational budgets. Moreover, ELPPM has a powerful ability to capture multi-scale context using pyramid pooling architecture. Finally, the boundary supervision provides additional cues to boost performance. The experimental results show DRBANet achieves available trade-off in terms of segmentation accuracy and implementing efficiency. In the future, we are interested in transferring our model to the tasks of object detection \cite{detection1,detection3} and image classification \cite{hrnet,Shufflenetv2}.

\bibliographystyle{ieeetr}
\bibliography{refs1.bib}
\end{document}